Graphical Abstract

**Artificial Intelligence based tool wear and defect prediction for special purpose milling machinery using low-cost acceleration sensor retrofits**

Mahmoud Kheir-Eddine,Michael Banf,Gregor Steinhagen

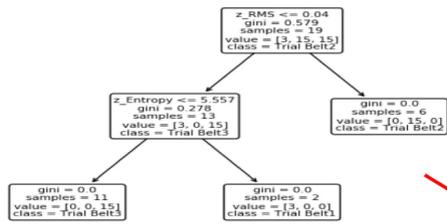

Classification

Wavelet analysis

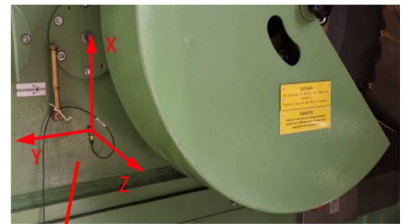

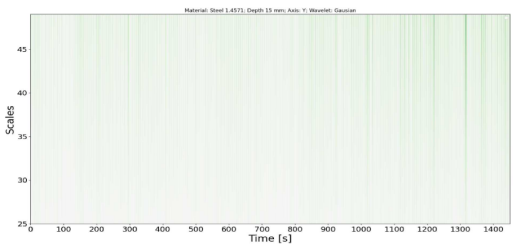

Acceleration Signal

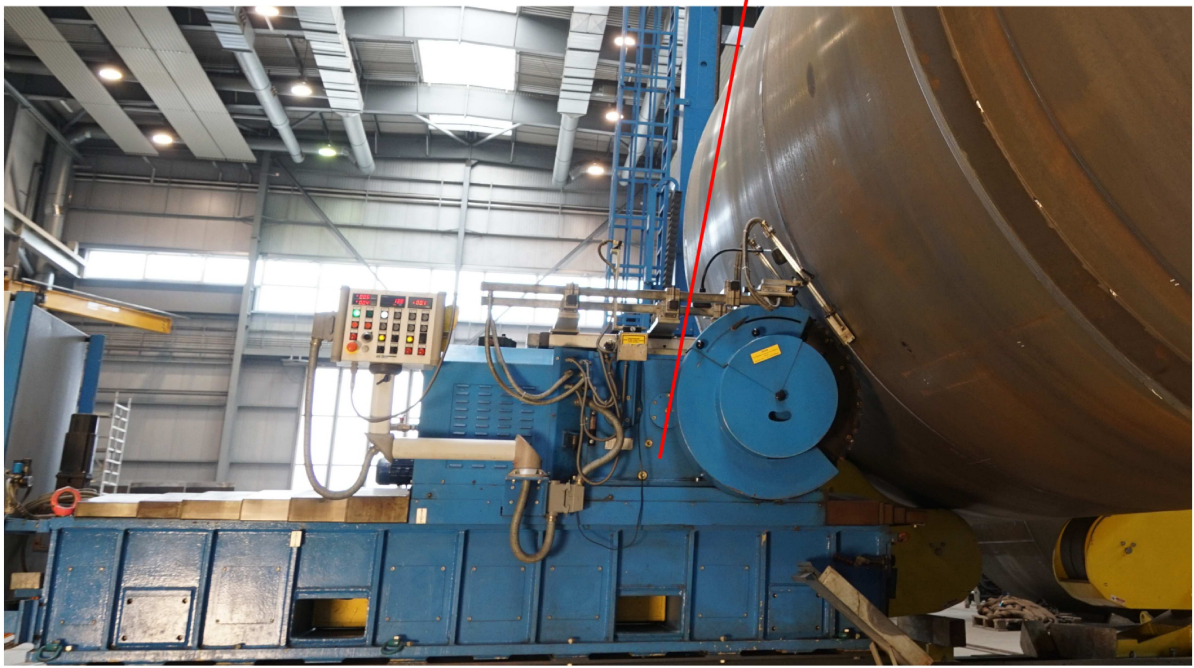

Highlights

**Artificial Intelligence based tool wear and defect prediction for special purpose milling machinery using low-cost acceleration sensor retrofits**

Mahmoud Kheir-Eddine,Michael Banf,Gregor Steinhagen

- A special purpose milling machine is equipped with an acceleration sensor in order to collect experimental data for multiple machine conditions.

- Condition specific experimental data is analyzed using continuous wavelet transforms to visualize effects and identify relevant frequency ranges.

- Different features extraction strategies are combined with multiple supervised classification frameworks to evaluate the feasibility of milling tool anomaly and defect detection from acceleration sensor signal data.

# Artificial Intelligence based tool wear and defect prediction for special purpose milling machinery using low-cost acceleration sensor retrofits

Mahmoud Kheir-Eddine[a,b], Michael Banf[a] and Gregor Steinhagen[a]

[a]fabforce GmbH & Co. KG., Nepthen, Germany
[b]TU Dortmund University, Germany

## ARTICLE INFO

## ABSTRACT

Milling machines form an integral part of many industrial processing chains. As a consequence, several machine learning based approaches for tool wear detection have been proposed in recent years, yet these methods mostly deal with standard milling machines, while machinery designed for more specialized tasks has gained only limited attention so far. This paper demonstrates the application of an acceleration sensor to allow for convenient condition monitoring of such a special purpose machine, i.e. round seam milling machine. We examine a variety of conditions including blade wear and blade breakage as well as improper machine mounting or insufficient transmission belt tension. In addition, we presents different approaches to supervised failure recognition with limited amounts of training data. Hence, aside theoretical insights, our analysis is of high, practical importance, since retrofitting older machines with acceleration sensors and an on-edge classification setup comes at low cost and effort, yet provides valuable insights into the state of the machine and tools in particular and the production process in general.

## 1. Introduction

Milling is a widely used process to produce workpiece contours. However, many aspects do affect the quality of the process, and, as a consequence, the produced workpieces. The vast majority of systems are CNC machines, but a number of special purpose milling machinery exists as well, employed e.g. for preparing weld seams and edges of large workpieces. To this end, special tool geometries are equipped with standard cutting blades in order to reduce operational costs. However, blade wear and degradation of merely a single blade, e.g. due to material failure or hitting inclusions in the base material, may lead to workpiece quality loss or even complete tool breakage. Therefore, this can cause delay and time consuming manual effort while rejected products accumulate wasted upstream factory capacity, material, labor and cost. Hence, early detection of tool wear is becoming invaluable to reduce maintenance costs as well as material or product waste, and various approaches based on different sensor concepts and analysis strategies have been proposed during recent years Wang et al. (2014); García-Nieto et al. (2016); Zhang et al. (2017); Hesser and Markert (2019). We refer to Mohanraj et al. (2020) for a comprehensive overview of current research.

Here, we describe the application of an acceleration sensor to a highly task-specific, round-seam milling machine, used during the fabrication of large pipe and tubular structures. Although tool wear analyses, e.g. as described in Wang et al. (2014); Zhang et al. (2017) or Hesser and Markert (2019), also utilize vibration data, their work centers around the study of standard milling machines, which have been subject of extensive research already. In contrast, our study concentrates on a special type of milling machine comprising of larger and more complex tool geometries with an irregular blade pitch. In addition, while the aforementioned studies focus on homogeneous wear of cutting blades, we also consider degradation or sudden defects of individual blades.

Furthermore, we examine effects regarding machine stability, including machine mounting as well as tension of the toothed belt connecting the electrical motor to the gearbox. Since machine stability significantly contributes to process quality, it is important to get an early feedback regarding any instabilities.

The tooth belt tensions and its maintenance typically requires a manual measurement on an inactive machine. An automatic in-process detection may lead to further optimizations in the maintenance schedule. To this end, our paper also evaluates different approaches to failure recognition. Hence, aside theoretical findings, our analysis is of high, practical importance, since retrofitting older machines, such as the round-seam milling machine, with a single or a set of vibration sensors comes at low cost and effort, yet provides valuable insights into the state of the machine and tools in particular and the production process in general.

In our study, instead of using an end-to-end learning based approach to tool wear detection, as e.g. in Martínez-Arellano et al. (2019); Wang et al. (2020), we resort to a classical separation of feature analysis and extraction followed by a simple, supervised classification setup. This is in part to allow for an initial analysis of a minimal set of features, but also to evaluate on-edge computation capable classification frameworks, in particular in the context of very limited training data. We experiment with two feature design strategies, i.e. either extracting basic statistical features directly computed on the raw vibration signal inputs as well as an automated pre-processing scheme via

m.banf@fabforce.com (M. Banf); g.steinhagen@fabforce.com (G. Steinhagen)
ORCID(s):

Wavelet Packet Decomposition Mallat (1999). Taking one-edge computability as well as small amounts of training data into consideration, we resort to Support Vector Machines Cortes and Vapnik (1995) and Random Forests Breiman (2001) classification frameworks for our analysis and show that different approaches are advantageous for individual condition monitoring aspects of the machine.

The remaining manuscript is structured as follows. Section 2 will give an introduction to the round seam milling machine, while section 3 describes our two-step strategy of data pre-analysis and feature based classification. Section 4.3 describes the approach to detect blade wear while section 4.2 discusses the detection of distinct defects related to individual sets of blades. Section 4.3 shows the results regarding machine stability and section 4.4 regarding the tooth belt tension.

## 2. Machine and experimental design

The machine that we focus on in this paper is a round-seam milling machine (fig. 1) manufactured by Graebener Maschinentechnik GmbH & Co. KG in Germany. Its applications include the preparation of weld seams on steel tubular structures, e.g. in wind power towers or pressure vessel production. After the inner side of a tube has been welded, the outer edges are milled in order to yield a proper welding geometry.

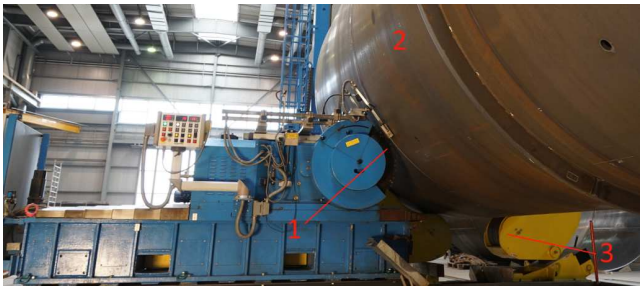

**Figure 1**: Example of the Graebener round-seam milling machine: The milling tool (1) is positioned in front of a workpiece (2), with the workpiece itself being located on a turning device (3) to control material feed throughout the milling process.

Therefore, a human operator positions the machine in front of the tube structure, while the tube is mounted on four rotating wheels. These supporting wheels allow for the tube to be rotated in a down milling process that is the cutting tool is fed in direction of its rotation. Moreover, the milling spindle position along its axis of rotation is adjustable in order to react to any geometric distortions such as out-of-roundness of the tube. Different milling tools and hard metal cutting blades may be mounted onto the milling tool, resulting in a variety of possible tool layouts regarding differences in diameter, opening angle and opening radius.

Despite these tool layout differences, its general setup and purpose always remains the same. It is a combination of circular blades performing the main material removal as well as lateral blades that widen the cut for deeper seam preparations. Figure 2 shows the milling tool of the used machine. For our experiments, it was equipped with a 700 mm diameter tool holding 32 circular blades and 24 lateral blades. These circular blades cover half the milling cut geometry which is defined by a radius and an opening angle. The angle distance between individual blades is not equally distributed to reduce induced vibrations.

In order to achieve a final weld seam geometry, the milling process is usually performed in different phases. First, while being rotated, the first intermediate geometry is cut starting at the workpiece's surface. It has a predefined depth which does not need to be the final depth. After each rotation of the workpiece, the milling depth is adjusted. Depending on the desired depth of the final geometry as well as the material parameters, the number of phases and their depth differences are adapted.

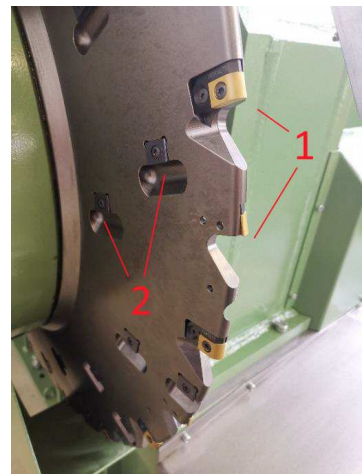

**Figure 2**: Milling tool with circular (1) and lateral (2) blades.

Compared to standard CNC milling machines, the round-seam milling machine's tool, given its high number of cutting blades on a large diameter, rotates with a lower frequency up to around 101 rpm. In this manner, regular cutting speeds and a high material removal can be achieved with a comparably low rotation frequency.

The machine's power train consists of a gear box, an electric motor as well as a toothed belt between motor and gear box. This transmission belt needs to be fastened on a regular basis. The power of the motor in our experiments was 30 KW. Given these machine specifications, four types of defects are of particular interest for further analysis:

*Tool wear of the blades* The blade inserts become blunt after a certain amount of milling depending on material and cutting conditions. Further, tool wear affects the milling quality and the risk of a blade breakage. An optimized replacement strategy will ensure quality and reduce costs. (Section 4.3).

*Defect of individual and groups of blades* Individual blades might fail due to material weaknesses in the blades themselves or by a blade hitting material impurities such as weld seam powder still remaining on the surface of the workpiece. Failure of single blades, if undetected, can cause great damage not only to the workpiece but also the milling tool or machine (Section 4.2).

*test* Since the operator positions the machine in front of the workpiece manually, damages to the workpiece or the machine itself may arise from improper machine stability, e.g due to an uneven shop floor, an improperly adjusted machine, or the machine being placed on obstacles of any kind (Section 4.3).

*Tooth belt fixture* The belt in the power train looses tension over time and needs a regular, manual, fastening. Early detection of a its loosening and informing a human operator about the belt's tension declining below some critial level can prevent undesired problems when milling a workpiece (Section 4.4).

We will address all these topics in subsequent sections. For our experiments we use a Brüel and Kjær triaxial acceleration sensor type 4535-B with a measuring band from 0.3 Hz to 10 kHz. After checking several locations for placing the device in a preliminary analysis, the position directly at the gear box appeared to be the most promising. Figure 3 shows the position and the corresponding coordinate system for the acceleration signals. We acquired the data with LabView and a National Instruments NI 9230 BNC DAQ amplifier and used the recorded files for our analysis.

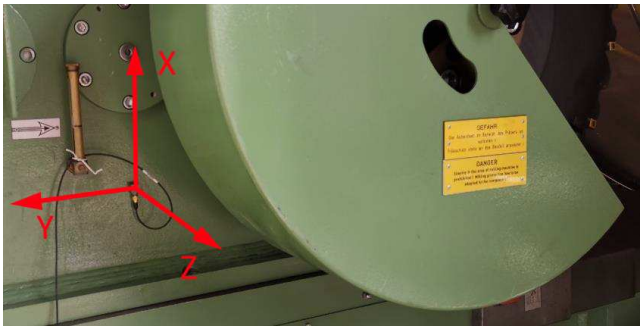

**Figure 3**: Position of the acceleration sensor with corresponding coordinate system. All axes of rotation of the different rotating elements in the machine are along the Z-axis of the sensor.

Since we mount the sensor to an external position, no changes to the machine were necessary. This makes our approach highly convenient to allow for low-cost retrofits of existing machinery which is of interest given the widespread availability of similar round-seam milling machines being in use for a long time already.

## 3. Two-step data analysis strategy

First, the combination of our employed sensor, rotation frequency as well as the number of blades used allows for a good resolution of single blade impacts. We performed a series of continuous wavelet transforms Mallat (1999) based analyses to compare the effects of the different, aforementioned types of machinery issues. As we discuss in section 4.3 this wavelet transformations and subsequent time-dependent frequency spectrum visualization allow for a high resolution, temporal overview of even single blade impacts on the material, as well as the detection of distinct events in the milling process. Furthermore, it helps elucidate relevant acceleration axes and frequency ranges.

Next, we devise several defect classification experiments deriving features sets either based on wavelet package decomposition as in Wang et al. (2014), or selecting statistical features directly estimated from the raw sensor signals as in Knittel et al. (2019); García-Nieto et al. (2016). We then compare performances of two types of classification frameworks, i.e. Random Forest and Support Vector Machines, given these feature extractions. Support Vector Machines have previously been used in the context of milling tool wear detection e.g. by García-Nieto et al. (2016); Wang et al. (2014); Knittel et al. (2019).

## 4. Experiments and results

### 4.1. Blade wear

In order to analyze and detect blade wear, we measured the milling in a real production process of a 1.4571 stainless steel which is used for pressure vessel construction. The milling of the entire seam preparation took 24.8 minutes for the first phase. Since the material is very tough, this already caused considerable blade wear. The milling tool rotated at 41 rpm which corresponds to a cutting speed of about 1.5 m/s, tooth pass frequency of 21.86 Hz for the circular blades. We set the material feed velocity of 340 mm/min. The tooth pass frequency in general is not an harmonic frequency because of the mentioned unequal distribution of the circular blades. The depth of the cut was 15 mm. All data was recorded with a sampling frequency of 6.4 kHz.

To get an overview of the data we first applied a continuous wavelet transform using a Gaussian wavelets as basis function Mallat (1999). Figure 4 highlights the resulting wavelet transformation of the recorded 24.8 minutes milling within relevant scaling ranges between 25 and 50 for the Y-axis signal of the acceleration. As can be observed, even individual hits of the blades on the milling surface are visible. Further, worn blades cause a higher signal response, resulting in a noticeable difference between the initial phase, with newly replaced blades, and the end of the milling trial, after significant wear has occurred. Finally, we notice a phase of weaker impacts in the middle of the trial, possibly due to significant additional effects during production, e.g. caused by the bearing of the workpiece, the alignment of the milling machine or the precision of the workpiece itself. However, these effects do compare to machine stability related effects described in section 4.3. Since milling had been recorded in a real production process, the collected data can be seen as representative for other production scenarios.

Following this analysis and visualization of the input signals, we devise a classification framework for blade wear. For feature extraction, we use Wavelet Packet Decomposition Mallat (1999), an approximation of the discrete wavelet transform based on filter banks, applied to both the approximation coefficients and the detail coefficients. For a more in-depth review of Wavelet Packet Decomposition we refer to Gokhale and Khanduja (2010).

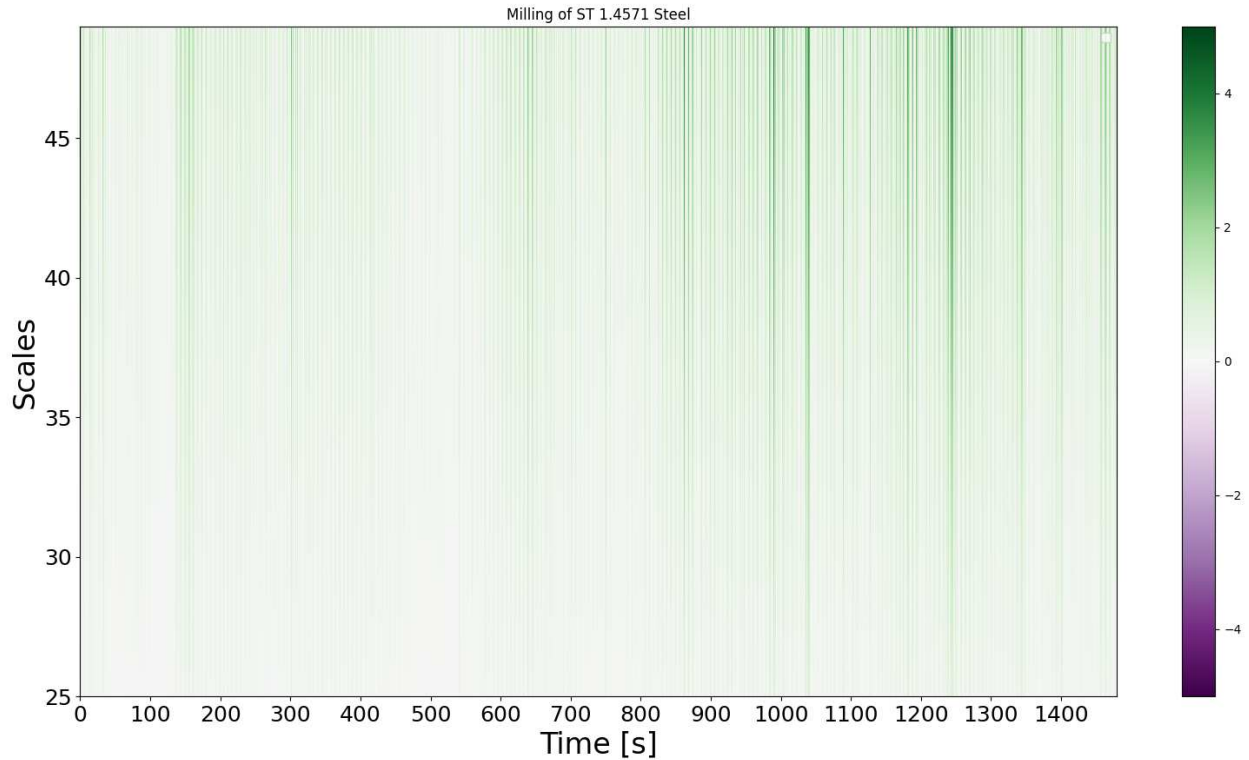

Decomposition level is set to three and the root mean square values of the resulting sub bands are used for feature extraction. To establish a reasonable dataset for classification model training and avoid ambiguities in the training data itself, we divide the recorded set of signals into two groups of non-overlapping, one second long samples; the non-defect (i.e. fresh blades) class being samples collected from the first five minutes of the trial, and the defect (i.e. worn blades) class samples taken from the last five minutes. This results in a training set of 300 samples per class. Following the creation of a training set, we train two generic classification frameworks, i.e. Random Forest and Support Vector Machines. Model training is performed with a validation split of 25 percent on the training dataset. The classifiers achieved accuracies of 98 percent (Support Vector Machine) and 95 percent (Random Forest). A feature importance analysis based on the trained Random Forest classifier revealed the sub band between 2 and 2.4 kHz to rank highest, providing valuable insight into necessary measuring ranges of a sensor for milling with 41 rpm. Further, these model performances demonstrate the feasibility of tool wear detection. However, since a set of blades is usually used in different material and machine setting combinations, these aspects are subject to further investigation.

## 4.2. Defect of individual blades

As discussed, the breaking or blunting of individual blades may cause considerable problems in round-seam milling. We therefore designed an experiment in order to simulate such an event occurring. We first used a fresh new set of blades to establish a baseline for our comparative analysis. Then we successively inserted two and six worn blades in two individual experiments. These were evenly distorted on the blades positions. With each setup we milled a steel tube of 1.0045 material. The milling tool rotated with 101 rpm which resulted in a tool pass frequency of 53.87 Hz and a cutting speed of 3.7 m/s. The feed rate was 500 mm/min and the depth of the cut 14 mm. However, in contrast to our first experiment this time, we measured the second phase of milling, i.e. deepening a cut from a 19 mm deep geometry to a 33 mm deep one.

Figure 5 shows three Morlet Wavelet based wavelet transforms of the accelerometer's Y-axis signal for the three experimental conditions, i.e. normal (i.e. new), two worn and six worn blades. As opposed to entirely new blades being used, milling with two worn blades exhibits significant differences in the scales around 30. This difference only increases with six inserts of worn blades revealing higher frequency amplitudes around the scale of six.

In this experiment, we create a three-class classification framework for the three experimental conditions, dividing recorded data in non-overlapping samples of one second length. We now derive a feature set per training sample using some well-documented statistical features, directly as in previous experiments but loosening the mounting of the machine. The milling settings were the same as before except for the milling phase. This time the initial phase of 19 mm was studied since the machine stability should be recognized in the beginning of a milling process.

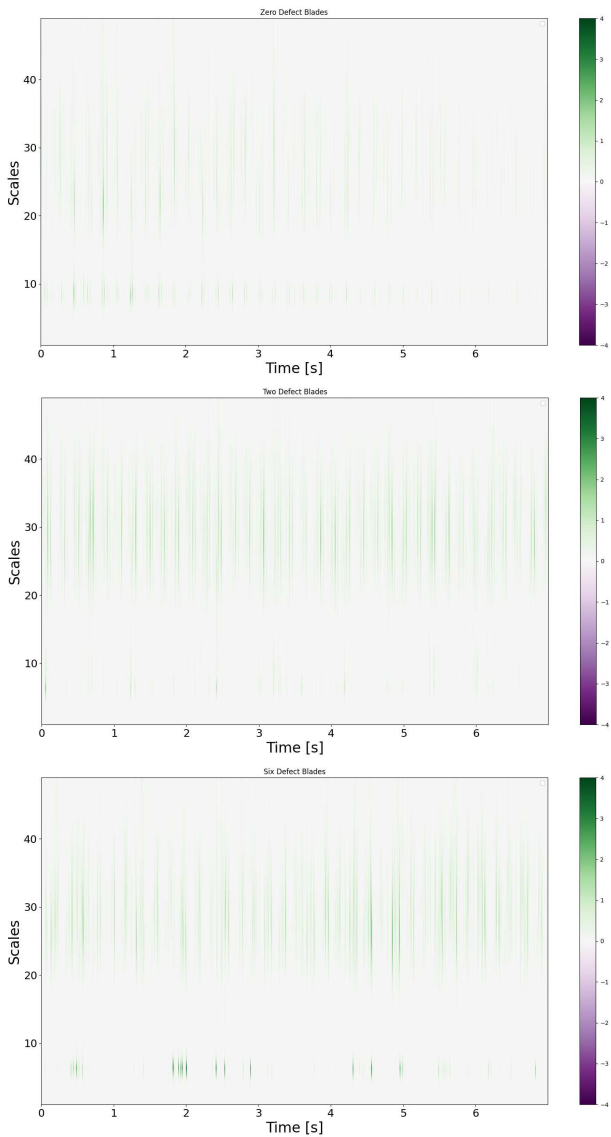

**Figure 5**: Morlet Wavelet based wavelet transforms of the accelerometer's Y-axis signal for three experimental condition, i.e. normal (top panel), two (middle panel) and six worn (bottom panel) blades.

Again we perform a comparative analysis of respective Morlet wavelet based wavelet transforms of the recorded signals with a stable as well as an unstable machine mount. Figure 6 shows the analysis for the accelerometer's Y-axis signal. One can see that an unstable machine causes more disturbances in the lower scales around six.

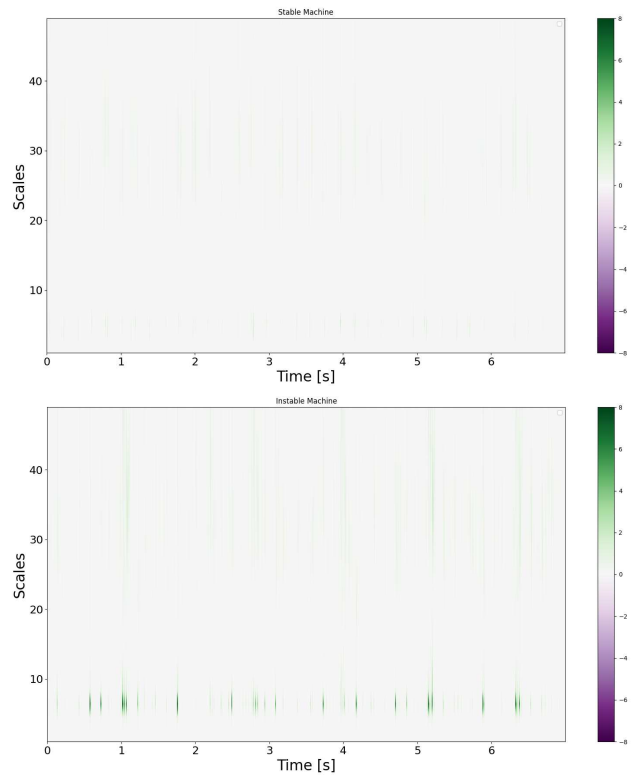

**Figure 6**: Morlet Wavelet based wavelet transforms of the accelerometer's Y-axis signal for two experimental condition, i.e. stable machine mounting (top panel) and unstable machine mounting (bottom panel)

estimated from the raw input signal, including root mean square, variance, entropy, shape factor, crest factor, kurtosis, skewness, minimum and maximum. For a detailed overview of these features we refer to Caesarendra and Tjahjowidodo (2017). Model training of the two classifiers is again performed with a validation split of 25 percent on the resulting training set, with both classifiers achieving accuracies of 100 percent (Support Vector Machine) and 94 percent (Random Forest). These results again encourage the use of a lightweight, in-process machine monitoring framework to inform an operator about putative process instabilities.

### 4.3. Machine stability

In order to analyse the results of insufficient machine stability we performed a separate trial, milling the same tube

Using the same input sensor signal subdivision strategy and validation splits as described combined with the statistical feature extraction scheme as in section 4.2, we achieve classifier accuracies of 100 percent (Support Vector Machine) and 95 percent (Random Forest). Note that the only difference to afore-mentioned classification setups is a smaller training dataset given that an unstable machine causes considerable stress for the machine, limiting the duration of the trial. Furthermore, it is important to be able to detect machine stability issues in only a short period of time, constituting another aspect of further inquiry.

### 4.4. Tooth belt fixture

The last aspect of our analysis is the effect of tooth belt fixture. Since the milling process induces very high frequencies in the machine, we want to monitor toothed belt fixture independently and in an idle machine. This can be tested in the machine setup phase before milling to ensure the toothed belt to be fastened properly.

Therefore, the machine was run in idle mode with 101 rpm in three toothed belt settings. We ran three trials, using a correctly adjusted, a partially and an extremely loosened belt. Figure 7 shows the results of a wavelet transformation of the accelerometer's Z-axis signal for the three conditions using Shannon Wavelets as basis function.

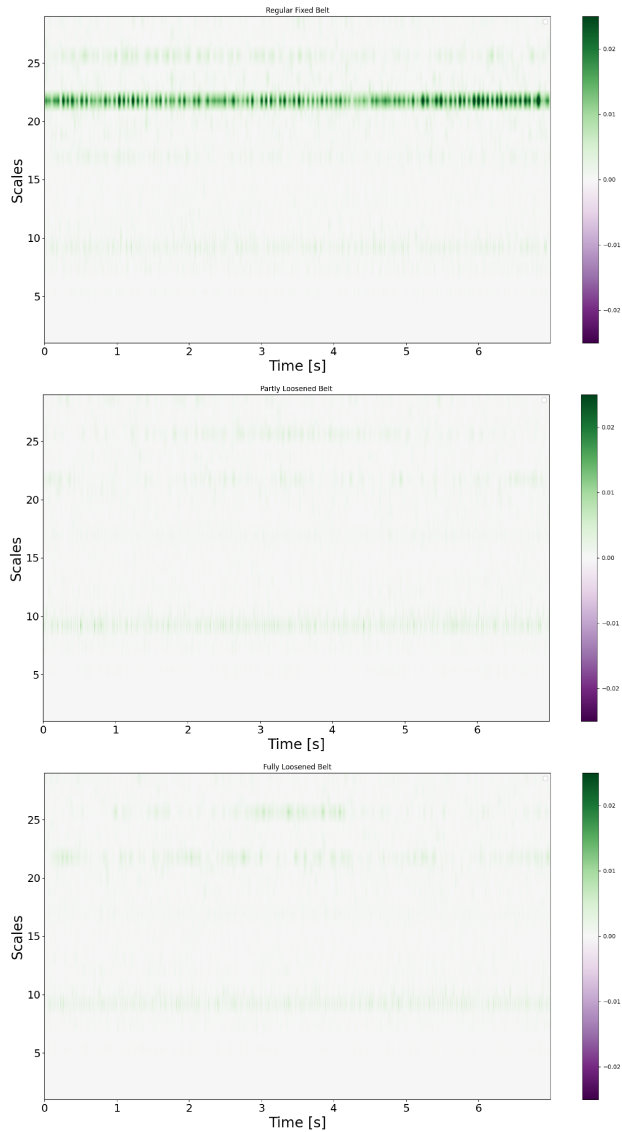

**Figure 7:** Shannon Wavelet based wavelet transforms of the accelerometer's Z-axis signal for three experimental condition, i.e. fixed (top panel), partly loosened (middle panel) and extremely loosened (bottom panel) belt.

Since the machine was run in idle mode, the signal strength is much lower and both Morlet or Gaussian Wavelets do not reveal the differences between experiments. Hence, we employ Shannon Wavelets Mallat (1999) and adapt the wavelet bandwidth to 0.1 and the center frequency to 3 in order to visualize the differences. The fact that the accelerometer's Z-axis signal exhibits the main differences between conditions may be explained by the skew gear mounted on the same axis of the transmission pulley holding the toothed belt, which leads to an excitation in the Z-Axis. In this comparative approach we achieved a clear and visible difference between a regular belt fixation and the partly loosened toothed belt. They are visible in the scales around 22. The wavelet analysis plots a pure and continuous frequency of the well adjusted toothed belt. This frequency can be barely plotted with a partially loosened toothed belt, whereas it presents a discontinuous and erratic behavior in the case of an extremely loosened toothed belt.

Using the same input sensor signal subdivision strategy as well as similar validation splits as described section 4.2 and using the statistical feature extraction scheme based on the accelerometer's Z-axis signal, we achieve classifier accuracies of 91 percent (Support Vector Machine) and 100 percent (Random Forest) on a very small dataset of 45 samples in total, each of one second length. This is encouraging, as it indicates that in-process classification is possible, even in the (realistic) case of highly limited data to train on.

## 5. Conclusion

Overall the results of our analyses are promising. In particular the tooth belt fixture and machine stability surveillance results can be directly translated into a machine control extension. We also show that different wavelet and machine learning approaches are suitable for different aspects of the analysis. In order to tackle these aspects, a flexible framework which compares different approaches and applies the best ones on an edge device would be of practical interest. A possible architecture in a broader production setup is shown in Steinhagen and Banf (2021).

Our analysis further reveals the importance of individual sensor axis recordings for different conditions. While Y-axis recordings seem more relevant for blade wear, blade defects and machine stability analysis, tooth belt stability could be best detected using Z-axis signals. Aside such theoretical findings, our analysis also is of high, practical importance, since retrofitting older machines with acceleration sensors and on-edge classification setups comes at low cost and effort, yet provides valuable insights into the state of the machine and tools in particular and the production process in general. Hence, our results encourage the design of a simple retrofit kit for already existing machines. Similar machine setups such as edge milling machines in the ship building or pipeline and construction tube industry are an interesting further application.

In real production, difficulties in defect detection may arise regarding different, not yet handled circumstances such as material changes for the same set of blades, varying machine settings and machine positions. Given that wide ranges of materials have not been analyzed in the literature so far, more experiments and flexible regression based approaches are needed in order to achieve material independent results. We want to address this aspect in the future. Further, we are already working on approaches to supervise deployed machine learning models, in order to allow for unfamiliar workpiece material and machine setting combinations, which may lead to undefined classification results, to be identified and introduced into an existing machine learning model.

Furthermore, we will work on quantitative blade wear or tooth belt regression approaches, instead of purely qualitative classification methods, since the ability to estimate and forecast a tool's lifetime or the best time for belt maintenance are highly valuable insights for customers to achieve both quantity and quality in production without compromising one over the other.